\documentclass{article}

\usepackage{microtype}
\usepackage{graphicx}
\usepackage{booktabs} %

\usepackage{hyperref}

\usepackage[accepted]{icml2020}

\usepackage{graphicx}
\usepackage{subcaption}
\usepackage{amsmath}
\usepackage{amsthm}
\usepackage{amssymb}
\usepackage{bm}
\usepackage[mathscr]{euscript}
\usepackage{varwidth}
\usepackage{times}

\usepackage{cancel} 
\usepackage{bbm}
\usepackage{xspace}
\usepackage{stmaryrd}
\usepackage{enumitem}
\usepackage[dvipsnames]{xcolor}

\DeclareMathOperator{\sgn}{sgn}

\newtheorem{thm}{Theorem}[]

\newtheorem{mydef}[thm]{Definition}

\newtheorem{exa}[thm]{Example} 
 
\newtheorem{pro}[thm]{Proposition} 
\newenvironment{proof-sketch}{\noindent{\textit{Sketch of Proof.}}\hspace*{0em}}{\qed\bigskip}

\newcommand{\X}{\ensuremath{\mathbf{X}}}
\newcommand{\B}{\ensuremath{\mathbf{B}}}

\newcommand{\V}{\ensuremath{\mathbf{V}}}

\newcommand{\x}{\ensuremath{\boldsymbol{x}}}
\newcommand{\bo}{\ensuremath{\boldsymbol{b}}}

\newcommand{\formula}{\ensuremath{\Delta}\xspace}
\newcommand{\Va}{\mathcal{V}}
\newcommand{\R}{\ensuremath{\mathbb{R}}\xspace}
\newcommand{\Bool}{\ensuremath{\mathbb{B}}\xspace}
\newcommand{\ch}{\ensuremath{\mathsf{ch}}}
\newcommand{\pa}{\ensuremath{\mathsf{pa}}}

\newcommand{\prob}{\ensuremath{\mathsf{Pr}}}

\newcommand{\theory}{\ensuremath{\Delta}}
\newcommand{\lra}{\mathcal{LRA}}

\newcommand{\bigO}{\mathcal{O}}
\newcommand{\WMI}{\ensuremath{\mathsf{WMI}}\xspace}

\newcommand{\WeightMI}[1]{\ensuremath{\mathsf{WMI}(#1)}\xspace}
\newcommand{\twoC}{\ensuremath{\mathsf{2}}\xspace}
\newcommand{\wfamily}[1]{\ensuremath{\boldsymbol{\Omega}^{#1}}\xspace}
\newcommand{\MI}{\ensuremath{\mathsf{MI}}}

\newcommand{\graph}{\mathcal{G}}
\newcommand{\primalgraph}{\mathcal{G}}

\newcommand{\smt}{SMT\xspace}
\newcommand{\smtlra}{SMT($\lra$)\xspace}

\newcommand{\weights}{w}
\newcommand{\factor}[1]{f_{#1}}
\newcommand{\clique}{\mathcal{S}}
\newcommand{\Clause}{\mathit{CLS}}
\newcommand{\Literal}{\mathit{LITS}}
\newcommand{\vars}{\ensuremath{\mathsf{vars}}}
\newcommand{\id}[1]{\llbracket{#1}\rrbracket}

\newcommand{\msg}[3]{\mathsf{m}_{{#1} \rightarrow {#2}}^{#3}}
\newcommand{\neigh}{\ensuremath{\mathsf{neigh}}}

\newcommand{\up}{+}
\newcommand{\down}{-}

\newcommand{\mathL}{\mathcal{L}}

\newcommand{\E}{\mathcal{E}}
\newcommand{\F}{\mathcal{F}}

\newcommand{\true}[0]{\texttt{true}}
\newcommand{\false}[0]{\texttt{false}}

\newcommand{\perf}{X}
\newcommand{\team}{\mathcal{T}}
\newcommand{\squad}{\mathit{B}}
\newcommand{\tree}{\mathsf{tree}}

\mathchardef\mhyphen="2D

\newenvironment{shrinkeq}[1]
{\bgroup
\addtolength\abovedisplayshortskip{#1}
\addtolength\abovedisplayskip{#1}
\addtolength\belowdisplayshortskip{#1}
\addtolength\belowdisplayskip{#1}}
{\egroup\ignorespacesafterend}

\usepackage{tikz}
\usetikzlibrary{positioning}
\usetikzlibrary{shapes}
\usetikzlibrary{fit}
\usetikzlibrary{chains}
\usetikzlibrary{arrows}
\usetikzlibrary{calc}

\newcommand{\midlinewidth}{1.0pt}
\newcommand{\incmidlinewidth}{1.5pt}

\newcommand{\sqboxs}{1.2ex}%
\newcommand{\sqbox}[1]{\textcolor{#1}{\rule{\sqboxs}{\sqboxs}}}
\definecolor{cblue}{RGB}{54,91,183}
\definecolor{cred}{RGB}{241,133,136}
\definecolor{cgreen}{RGB}{168,208,151}

\newcommand{\yscale}{0.18}
\newcommand{\xscale}{0.7}
\newcommand{\axisoffset}{8pt}
\newcommand{\figoffset}{2}
\newcommand{\ymax}{1.1}

\definecolor{petroil2} {RGB} {36, 165, 175}

\icmltitlerunning{Scaling up Hybrid Probabilistic Inference with Logical and Arithmetic Constraints via Message Passing}

\begin{document}

\twocolumn[
\icmltitle{Scaling up Hybrid Probabilistic Inference \\ with Logical and Arithmetic Constraints via Message Passing}

\icmlsetsymbol{equal}{*}

\begin{icmlauthorlist}
\icmlauthor{Zhe Zeng}{equal,ucla}
\icmlauthor{Paolo Morettin}{equal,trento}
\icmlauthor{Fanqi Yan}{equal,amss}
\icmlauthor{Antonio Vergari}{ucla}
\icmlauthor{Guy Van den Broeck}{ucla}
\end{icmlauthorlist}

\icmlaffiliation{ucla}{Computer Science Department
  University of California, Los Angeles, USA}
\icmlaffiliation{trento}{DISI, University of Trento, Italy, Italy}
\icmlaffiliation{amss}{Academy of Mathematics and Systems Science, Chinese Academy of Sciences, China}

\icmlcorrespondingauthor{Zhe Zeng}{zhezeng@cs.ucla.edu}

\icmlkeywords{Probabilistic Inference, SMT, WMI}

\vskip 0.3in
]

\printAffiliationsAndNotice{\icmlEqualContribution} %

\begin{abstract}
Weighted model integration (WMI) is an appealing framework for probabilistic inference: it allows for expressing the complex dependencies in real-world problems, where variables are both continuous and discrete, via the language of Satisfiability Modulo Theories (SMT), as well as to compute probabilistic queries with complex logical and arithmetic constraints.
Yet, existing WMI solvers are not ready to scale to these problems. 
They either ignore the intrinsic dependency structure of the problem entirely,
or they are limited to overly restrictive structures.
To narrow this gap, 
we derive a factorized WMI computation enabling us to devise a scalable WMI solver based on message passing, called MP-WMI.
Namely, MP-WMI is the first WMI solver that can
(i)~perform exact inference on the full class of tree-structured WMI problems, and
(ii)~perform inter-query amortization, e.g., to compute all marginal densities simultaneously.
Experimental results show that our solver dramatically outperforms the existing WMI solvers on a large set of benchmarks.

\end{abstract}

\section{Introduction}
\label{sec:introduction}

In many real-world scenarios, performing probabilistic inference requires reasoning 
over domains with complex logical and arithmetic constraints while dealing with variables that are heterogeneous in nature, i.e., both continuous and discrete.
Consider for example the task of matching players in a game by their skills.
Performing probabilistic inference for this task has been popularized by~\citet{minka2018trueskill} and is at the core of several online gaming services.
A probabilistic model for this task has to deal with continuous variables, such as the player and team performance,
and reason over discrete attributes such as membership in a squad and the achieved scores.
Moreover, such a model would need to take into account constraints such as the team performance being bounded by that of the players in it, and that forming a squad boosts performance.
Ultimately, this translates into performing probabilistic inference in the presence of logical and arithmetic constraints and dependencies.

These hybrid scenarios are beyond the reach of probabilistic models including variational autoencoders~\cite{kingma2013auto} and generative adversarial networks~\cite{goodfellow2014generative}, whose inference capabilities, despite their recent success, are 
limited.
Classical probabilistic graphical models~\cite{koller2009probabilistic}, while providing more flexible inference routines, are generally incapacitated when dealing with continuous and discrete variables at once~\cite{shenoy2011inference}, or they tend to make simplistic~\cite{heckerman1995learning,lauritzen1989graphical} or overly strong assumptions about their parametric forms~\cite{yang2014mixed}.
Even recent efforts in modeling these 
hybrid scenarios while delivering tractable inference~\cite{molina2018mixed,vergari2019automatic} 
can not perform inference in the presence of complex constraints.

Weighted Model Integration (WMI)~\cite{belle2015probabilistic,morettin2017efficient} is a recent framework for probabilistic inference that offers all the aforementioned ``ingredients'' needed for hybrid probabilistic reasoning with logical constraints, \textit{by design}.
WMI leverages the expressive language of Satisfiability Modulo Theories (SMT)~\cite{barrett2010smt} for describing 
problems over continuous and discrete variables.
Moreover, WMI provides a principled way to perform hybrid probabilistic inference:
asking for the probability of a complex query with logical and arithmetic constraints can be done by integrating weight functions over the regions that satisfy the constraints and query at hand.

Despite these appealing features, current state-of-the-art WMI solvers are far from being applicable to high-dimensional real-world scenarios.
This is due to the fact that most solvers
ignore the
dependency structure
of the problem, here expressible through the notion of a primal or factor graph of an SMT formula~\cite{dechter2007and}.
Thus, their practical utility is limited 
by their inability to scale up the WMI inference.
In contrast,
SMI~\cite{zeng2019efficient}, is a recently proposed solver
that directly exploits the problem structure encoded in primal graphs
while reducing a WMI problem to an unweighted one.
However, in order to perform a tractable reduction, SMI is limited to a restricted set of weights, and hence a very narrow set of WMI problems.

The contribution we make in this work is twofold.
First, we theoretically trace the boundaries for the classes of tractable WMI problems %
known in the literature.
Second, we expand these boundaries by devising a polytime algorithm for exact WMI inference on a class that is strictly larger than the class previously known to be tractable.
Our proposed WMI solver, called MP-WMI, adopts a novel message-passing scheme for WMI problems.
It is able to exactly compute all the variable marginal densities
at once.
By doing so, we are able to scale inference beyond the capabilities of all current exact WMI solvers.
Moreover, we can amortize inference \textit{inter-queries} for rich SMT queries that conform to the problem structure.

The paper is organized as follows.
We start by reviewing the necessary SMT and WMI background.
Then we trace the boundaries between hard and tractable WMI problem classes in Section~\ref{sec: tractable}.
Next, we present our exact message-passing WMI solver in Section~\ref{sec: MP-WMI} together with its complexity analysis in Section~\ref{sec: complexity}.
Before comparing our solver to the existing WMI solvers on a set of benchmarks, we discuss related work in Section~\ref{sec: related work}.

\section{Background}
\label{sec:background}
\paragraph{Notation.} We use uppercase letters for random variables (e.g.,~$X,B$) and lowercase letters for their assignments (e.g.,~$x, b$). 
Bold uppercase letters denote sets of variables (e.g., $\X,\B$) and their lowercase denote their assignments (e.g., $\x,\bo$).
We represent logical formulas by capital Greek letters, (e.g.,~$\Lambda, \Phi,\Delta$), and literals (i.e., atomic formulas or their negation) by lowercase ones (e.g.,~$\phi,\delta$) or $\ell$.
We denote satisfaction of a formula $\Phi$ by one assignment $\x$ by $\x\models\Phi$ and we 
denote its corresponding indicator function as $\id{\x\models\Phi}$.
For undirected graphs, $\neigh$ denotes the set of neighboring nodes; for directed ones, $\pa$ and $\ch$ denote the parent node and the set of child nodes respectively.

\paragraph{Satisfiability Modulo Theories (SMT).} SMT~\cite{barrett2018satisfiability} generalizes the well-known SAT problem~\cite{biere2009handbook} to determining the satisfiability of a logical formula w.r.t.\  a decidable 
theory.
Rich mixed logical/arithmetic constraints can be expressed in SMT for hybrid domains.
In particular, we consider quantifier-free SMT formulas in the theory of linear arithmetic over the reals, or SMT($\lra$).
Here, formulas are propositional combinations of atomic Boolean literals
and of atomic $\lra$ literals over real variables, for which satisfaction
is defined in a natural way.
W.l.o.g. we assume SMT formulas to be in conjunctive normal form (CNF).
In the following, we will use the shorthand SMT to denote SMT($\lra$).

\begin{figure}[tb]
\centering
\begin{subfigure}[t]{0.25\textwidth}
    \centering
    \includegraphics[width=0.8\textwidth]{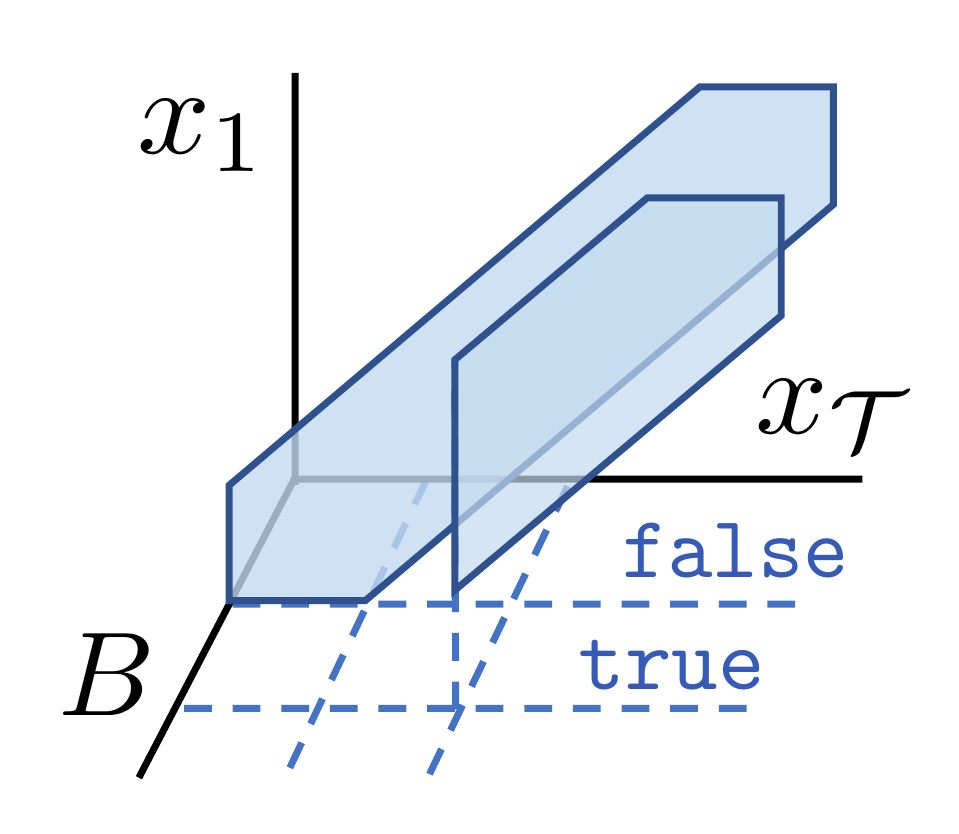}
\end{subfigure}
~
\begin{subfigure}[t]{0.2\textwidth}
\centering
    \begin{tikzpicture}[level/.style={sibling distance=15mm/#1, line width=\midlinewidth},level distance=9mm]
    \node [circle, draw, line width=\midlinewidth] (sq){$\squad$}
        child {node [circle, draw, scale=0.8, line width=\midlinewidth](pt){$X_\team$}
        child {node [circle, draw, scale=0.8, line width=\midlinewidth](x1){$X_1$}} 
        child {node [circle, draw, scale=0.8, line width=\midlinewidth](x2){$X_2$}} 
        child [missing]
        child {node [circle, draw, scale=0.8, line width=\midlinewidth](xn){$X_n$}}
        };
    \path (x2) -- (xn) node [midway] {$\cdots$};
    \end{tikzpicture}
\end{subfigure}
\caption{\textbf{Feasible region} (left) of formula $\Gamma$ with one player and \textbf{primal graph} (right) of formula $\Gamma$ with $n$ players from Example~\ref{exa: team performance SMT model}.
}
\label{fig: primal graph eg}
\end{figure}
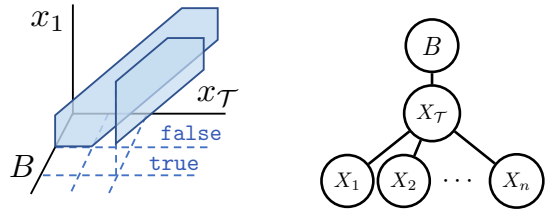

\begin{exa}[SMT representation of a skill matching system]
\label{exa: team performance SMT model}
In a skill rating system for online games,
the team performance $X_{\team}$ of each team $\team$ 
is defined by the performance $X_i$ of each player $i$ in team $\team$,
both of which are real variables.
The team performance $X_{\team}$ is also related to a Boolean variable $\squad$ 
indicating whether players in the team form a squad, i.e., a group of friends, which offsets (boosts) the team performance.
We can build an SMT formula $\Gamma$ of the relationship among these variables as follows.
For brevity, we omit the domains for real variables in the formula.
\begin{align*}
    \Gamma := \bigwedge_{i \in \team}~ \mid \perf_\team - \perf_i \mid < 1
        ~\bigwedge~ (\squad \Rightarrow \perf_\team > 2)
\end{align*}
We show in Figure~\ref{fig: primal graph eg} the feasible regions
of formula $\Gamma$
i.e., the volumes for which the constraints are satisfied.
\end{exa}

\paragraph{Weighted Model Integration (WMI).} 
Weighted Model Integration (WMI)~\cite{belle2015probabilistic,morettin2017efficient} provides a framework for probabilistic inference with models defined over the logical constraints given by SMT formulas.
\begin{mydef}(\textbf{WMI})
Let $\X$ be a set of continuous random variables defined over $\mathbb{R}$, 
and $\B$ a set of Boolean random variables defined over 
$\mathbb{B}=\{\true,\false\}$.
Given an SMT formula $\Delta$ 
over 
$\X$ and $\B$, and a \textit{weight function} 
$\weights:($\x$,$\bo$)\mapsto\mathbb{R}^{+}$
belonging to some parametric weight function family $\boldsymbol{\Omega}$,
the weighted model integration ($\WMI$) task computes
\begin{equation}
  \WMI(\theory, \weights;  \X, \B) \triangleq
  \sum\limits_{\bo \in\mathbb{B}^{|\B|}} 
  \int_{(\x, \bo) \models \formula} 
  \weights(\x, \bo) 
  \, d \x.
  \label{eq:wmi}
\end{equation}
\end{mydef}
That is, summing over all possible Boolean assignments $\bo \in\mathbb{B}^{|\B|}$ while integrating over the weighted assignments of $\X$ making the evaluation of the formula SAT: $(\x, \bo) \models \theory$.

Weight functions $\weights$ are usually defined as products of literal weights \citep{belle2015probabilistic,chavira2008probabilistic,zeng2019efficient}.
That is, for a set of literals $\mathL$, a set of per-literal weight functions $\mathcal{W} = \{ w_{\ell}(\x) \}_{\ell \in \mathL}$ is given,
with weight functions $w_{\ell}$ defined over variables in literal $\ell$.
Then, the weight of assignment $(\x, \bo)$ is:
\begin{equation*}
    \weights(\x, \bo) = \prod\nolimits_{\substack{\ell \in \mathL}} w_{\ell}(\x)^{\id{\x, \bo \models \ell}}. 
\end{equation*} 
When all variables are Boolean (i.e., $\X = \emptyset$), the per-literal weights $w_{\ell}(\x)$ are constants and we retrieve the original definition of the well-known weighted model counting~(WMC) task \citep{chavira2008probabilistic} as a special case of WMI.
In this paper, we assume that all per-literal weights are from some certain weight function family, and for literals not in the set $\mathL$, their weights are the constant function one.
This setting is expressive enough to approximate many continuous distributions 
\citep{belle2015probabilistic}.

\begin{exa}[WMI formulation of a skill matching system]
\label{exa: WMI formulation of a skill matching system}
Consider the team performance SMT model $\Gamma$ in Example~\ref{exa: team performance SMT model}.
Assume that a set of per-literal weights $w_{\ell_i}(\perf_\team, \perf_i) = 0.1 \cdot (\perf_\team + \perf_i - 6)^2$
is associated to literals $\ell_i = \perf_\team - \perf_i < 1$, quantifying how likely the team performance is upper bounded by player performances.
Then the \WMI of the formula $\Gamma$ with two players is $\WMI(\Gamma, \weights ; \X, \B) \approx 170.69$.
\end{exa}

Intuitively, $\WMI(\theory, \weights;  \X, \B)$ equals the partition function of the unnormalized probability distribution 
induced by weights $w$ on formula $\theory$.
In the following, we will adopt the shorthand $\WMI(\theory, \weights)$ for computing the WMI with all the variables in $\theory$ in scope.
The set of weight functions $\weights$ together act as an unnormalized probability density 
while the formula $\theory$ represents logical constraints defining its structure.
Therefore, it is possible to compute the (now normalized) probability of any logical \textit{query} $\Phi$ expressible as an SMT formula involving complex constraints
as
\begin{align*}
  \prob_{\formula}(\Phi) = \WMI(\formula\wedge\Phi, \weights) ~\slash ~ \WMI(\formula, \weights).  
\end{align*}

\begin{exa}[WMI inference for skill rating]
Suppose we want to quantify the squad effect in a 2v2 game.
Specifically, given two teams $\team_1$ and $\team_2$
whose players have the same performance, but team $\team_1$ is a squad while $\team_2$ is not,
that is, $\Phi_c = (\squad_1 = \true \land \squad_2 = \false)$.
We wonder what is the probability of query $\Phi = \perf_{\team_1} > \perf_{\team_2}$, that is team $\team_1$ wins and $\team_2$ loses.
The probability of query $\Phi$ can be computed by two \WMI tasks as follows.
\begin{align*}
    \prob_{\formula}(\Phi \!\mid\! \Phi_c) = \!\frac{\WMI(\formula \land \Phi_c \land \Phi, \weights)}{\WMI(\formula \land \Phi_c, \weights)}
    = \frac{4,206}{7,225} \approx 58.22 \%
\end{align*}
with the \smt formula $\formula := \Gamma_1 \land \Gamma_2$ where the two sub-formulas $\Gamma_1$ and $\Gamma_2$ model the two teams as in Example~\ref{exa: team performance SMT model}. 
\end{exa}

W.l.o.g, from here on we will focus on WMI problems on continuous variables only.
We can safely do this since a WMI problem on continuous and Boolean variables of the form
$\WMI(\theory, \weights;  \X, \B)$ can always be reduced in polytime to a new WMI problem $\WMI(\formula^\prime, \weights^\prime; \X^\prime)$ on continuous variables only, by properly introducing auxiliary variables in $\X^\prime$ to account for Boolean variables $\B$
without increasing the problem size~\cite{zeng2019efficient}.

\paragraph{From WMI to MI.}
Recently, model integration (MI) \cite{luu2014model} has been proposed as an alternative way to perform WMI inference in \citet{zeng2019efficient}.
MI is the task of computing the volumes corresponding to the models of an SMT formula.
As such, MI is a special case of WMI in which the weights equate to one everywhere.
\begin{mydef}{\textbf{(Model Integration)}}
Let $\X$ consist of continuous random variables over $\mathbb{R}$, 
and let $\theory$ be an SMT formula. The model integration (MI) of $\X$ over $\theory$ is:
\begin{equation*}
    \MI(\theory; \X) \triangleq \int_{\x \models \theory} 1 ~d \x 
    = \int_{\mathbb{R}^{|\X|}}\id{\x\models\theory} ~d\x 
    \label{eq:mi-vol-I}    
\end{equation*}
\end{mydef}
\citet{zeng2019efficient} propose a polytime reduction of a WMI problem with polynomial weights to 
an MI one such that their proposed MI solver is amenable to a certain class of WMI problems.
This reduction provides the basis for the largest class of tractable WMI problems known before our work.
We will review it in the next section,
before considerably expanding upon the class of WMI problems 
that can be solved tractably in the prior work.

\section{Tractable WMI inference}
\label{sec: tractable}

The major efforts in advancing WMI inference have been so far concentrated on devising sophisticated WMI solvers to deliver exact inference routines for general scenarios without investigating the effect of the structure of a WMI problems on its complexity. 
Little to no attention has gone to formally understand which classes of WMI problems can be guaranteed to be solved exactly {and in polynomial time,} that is, \textit{tractably}.

One notable exception can be found in~\citet{zeng2019efficient} where the search-based MI~(SMI) solver is introduced.
WMI problems for which SMI guarantees polytime exact inference constitute the first class of tractable WMI.
Intuitively, SMI solves MI problems by using search to leverage the conditional independence among variables.

As in~\citet{zeng2019efficient} we characterize the structure of an SMT formula via its \textit{primal graph}. 
\begin{mydef}{\textbf{(Primal graph of SMT)}}\label{def:smt primal}
  The primal graph
  of an 
  SMT formula $\formula$ is an undirected graph $\primalgraph_{\theory}$ whose vertices are variables in formula $\formula$ and whose edges connect any two variables that appear in a same clause in the formula $\formula$.
\end{mydef}

An example primal graph of the SMT formula in Example~\ref{exa: team performance SMT model} is shown in Figure~\ref{fig: primal graph eg}.
The SMI solver guarantees polynomial time execution for the class of MI problems with certain tree-shaped primal graphs, which we denote as $\tree\mathsf{MI}$.
\begin{mydef}{($\tree\mathsf{MI}$ Problem Class)}
\label{def:treeMI}
Let $\tree\mathsf{MI}$ be the set of all MI problems over real variables whose SMT formula $\formula$ induces a primal graph $\graph_{\theory}$ with treewidth one and with bounded diameter $d$.
Problems in $\tree\mathsf{MI}$ can be solved in polytime via SMI~\cite{zeng2019efficient}.
\end{mydef}{}

Note that in Definition~\ref{def:treeMI} the primal graph diameter here plays the role of a constant since, otherwise, SMI complexity can be worst-case exponential in diameter $d$.
In the following we will try to answer if larger classes than $\tree\mathsf{MI}$ are still amenable to tractable inference.
We start by demonstrating a novel result that states the hardness of a larger class of MI problems, still focusing on dependencies between two variables, but allowing for non-tree-shaped primal graphs.

\begin{mydef}{($\twoC\mathsf{MI}$ Problem Class)}
Let $\twoC\mathsf{MI}$ be the set of all MI problems over real variables whose SMT formula $\formula$ is 
a conjunction of clauses comprising at most two variables. 
\end{mydef}
Note that a clause comprising at most two variables can be a conjunction of
arbitrarily many literals.
Moreover, when there are more than two variables in a clause, in the primal graph there must be a loop and thus the treewidth of the primal graph is larger than one.
Hence all MI problems with tree-shaped primal graph must be in the class $\twoC\mathsf{MI}$.
\begin{figure}[!t]
    \centering
    \includegraphics[width=0.7\linewidth]{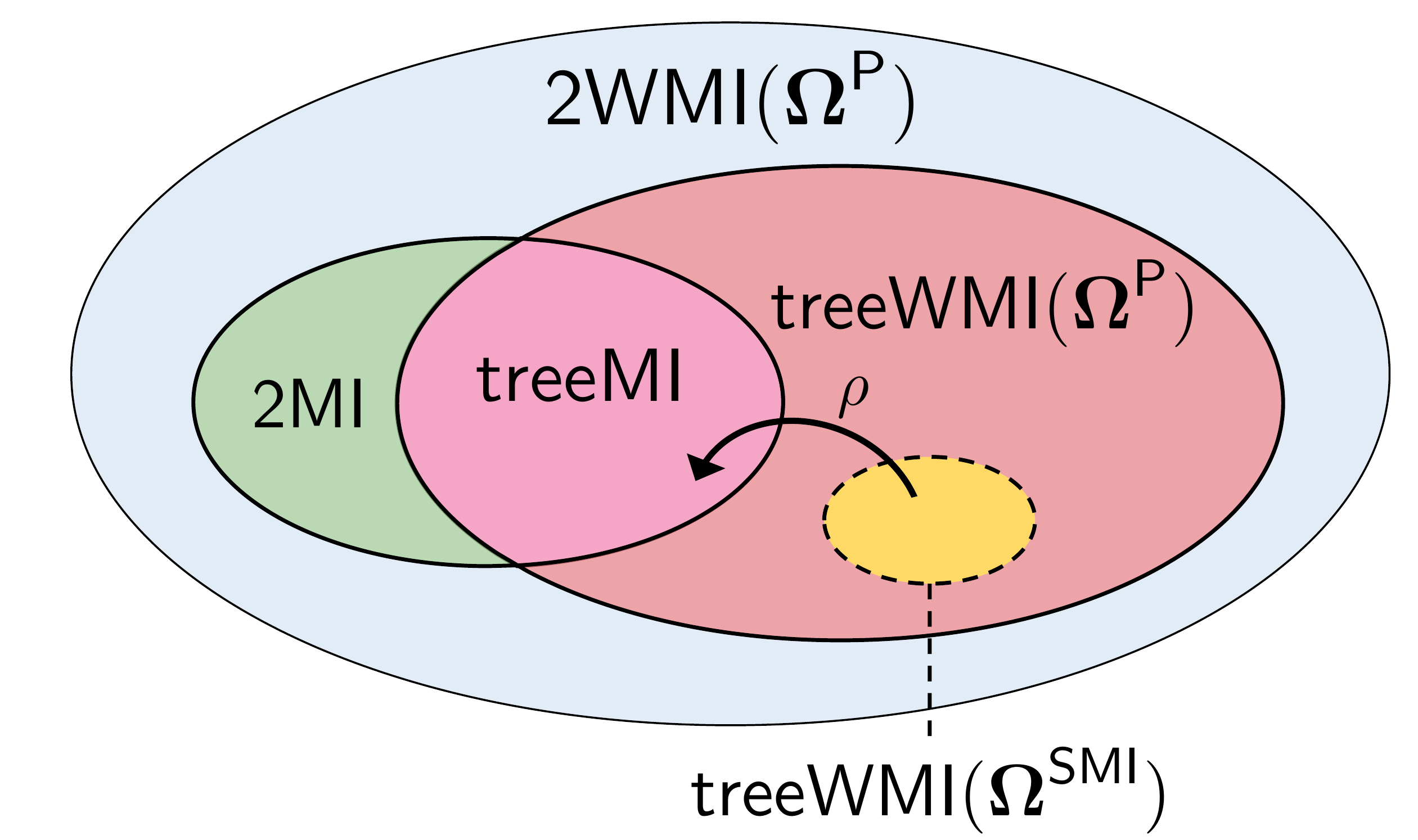}
    \caption{\textbf{The current landscape of classes of WMI problems.} 
    We enlarge the boundaries of tractable WMI inference from $\tree\MI$ to $\tree\WMI$ and prove the hardness of $\twoC\MI$ and $\twoC\WMI$.
    }
    \label{fig: wmi problem classes}
\end{figure}

\begin{thm}{(Hardness of $\twoC\mathsf{MI}$)}
\label{thm: sharp p hard of 2-Clause formula}
Given an MI problem in $\twoC\mathsf{MI}$ with an SMT formula $\formula$,
computing $\MI(\formula)$ is \#P-hard.
\end{thm}
\begin{proof-sketch}
The proof is done by reducing the \#P-complete problem \#2SAT to an MI problem in $\mathsf{\twoC\MI}$ with an SMT formula $\formula$ such that
counting the number of satisfying assignments to the \#2SAT problem equates to the MI of formula $\formula$. See Appendix for a detailed proof.
\end{proof-sketch}

From Theorem~\ref{thm: sharp p hard of 2-Clause formula} it follows that 
the problem class $\twoC\WeightMI{\wfamily{}}$, i.e., the WMI problems with SMT formulas being a conjunction of clauses comprising at most two variables, and with per-literal weights in weight function family $\wfamily{}$,
is also hard since class $\twoC\MI$ is a sub-class of $\twoC\WeightMI{\wfamily{}}$.
We revert our attention to WMI problems exhibiting a dependency tree structure.
Notice that for WMI problems, the tractability not only depends on the logical structure defined by the SMT formulas, but also the statistical structure defined by weight functions.
Next in our analysis, we take into consideration the weight function families.
Analogously to what Definition~\ref{def:treeMI} states, we introduce the notion of $\tree\WeightMI{\wfamily{}}$ with the associated weight function family specified as follows.
\begin{mydef}{($\tree\WeightMI{\wfamily{}}$ Problem Class)}
\label{def:treeWMI}
Let $\tree\WeightMI{\wfamily{}}$ be the set of all WMI problems over real variables whose SMT formula $\formula$ induces a primal graph $\graph_\formula$ with treewidth one and with bounded diameter $d$,
and whose per-literal weights are in a function family $\wfamily{}$.
\end{mydef}
\citet{zeng2019efficient} propose a WMI-to-MI reduction such that some $\tree\WeightMI{\wfamily{}}$ problems with polynomial weights are reduced in polynomial time to $\tree\MI$ problems amenable to tractable inference by the SMI solver.
Intuitively, the reduction process introduces auxiliary continuous variables and SMT formulas over these variables to encode the polynomial weight functions.
We refer the readers to \citet{zeng2019efficient} for a detailed description of the reduction. 
However, as shown next, the set of $\tree\WMI$ problems that can be reduced to $\tree\MI$ is rather limited.

\begin{mydef}{($\wfamily{\mathsf{SMI}}$ Weight Function Family)}
\label{def:OmegaMI}
Let $\wfamily{\mathsf{SMI}}$ be the family of per-literal weight functions that are monomials associated with either (i) univariate literals or 
(ii) a literal that appears exclusively in a unit clause, i.e., a clause consisting of a single literal.
\end{mydef}

\begin{thm}
\label{thm:rest-rho}
Let $\rho$ be the polytime WMI-to-MI reduction for $\tree\WeightMI{\wfamily{}}$ problems as defined in~\citet{zeng2019efficient}.
Then the image $\{ \rho(\nu) \mid \nu \in \tree\WeightMI{\wfamily{}} \} \subset \tree\MI$
if-and-only-if ~$\wfamily{} \subset \wfamily{\mathsf{SMI}}$.
\end{thm}
\begin{proof-sketch}
The necessary condition can be proved by the reduction process and the sufficient one can be proved by contradiction.
See Appendix for a detailed proof.
\end{proof-sketch}

Therefore, the SMI solver is limited to a rather restricted subset of $\tree\WeightMI{\wfamily{}}$
since from the definition of $\wfamily{\mathsf{SMI}}$ we can tell that it is a strict subset of monomial per-literal weights.
In order to enlarge the tractable class of WMI problems,
next we will define a rich family of weight functions.

\begin{mydef}{(Tractable Weight Conditions)}
\label{def:omegaMP}
Let $\wfamily{}$ be a family of per-literal weight functions.
We say that the tractable weight conditions~(TWC) hold for $\wfamily{}$ if we have:
\begin{enumerate}[label=(\roman*),noitemsep,topsep=0pt]
    \item closedness under product: $\forall f, g \in \wfamily{}$, $f \cdot g \in \wfamily{}$;
    \item tractable symbolic integration:
    $\forall f \in \wfamily{}$, the symbolic antiderivative of function $f$ can be tractably computed by symbolic integration; 
    \item closedness under definite integration:
    $\forall f \in \wfamily{}$ with its antiderivative denoted by $F$, given integration bounds $l(x), u(x)$ in $\lra$ with $x \in \X$, $F(u(x)) - F(l(x)) \in \wfamily{}$.
\end{enumerate}
\end{mydef}

Some example weight function families that satisfy TWC include the
polynomial family, exponentiated linear function family and the function family resulting from their product.
Moreover note that piecewise function families, when pieces belong to the above families, also satisfy TWC.
It turns out that the weight function families that satisfy TWC subsume and extend all the parametric weight functions adopted in the WMI literature so far.
The following proposition is a direct result
from the fact that the piecewise polynomial weight family $\wfamily{\mathsf{P}}$ is a strict superset of the family $\wfamily{\mathsf{SMI}}$.
\begin{pro}
Let $\wfamily{\mathsf{P}}$ be the piecewise polynomial weight function family.
The WMI problem class $\tree\WeightMI{\wfamily{\mathsf{P}}}$ is a strict superset of problem class $\tree\WeightMI{\wfamily{\mathsf{SMI}}}$.
\end{pro}
\begin{thm}
\label{thm:omegaMI-in-omegaMP}
If a weight function family $\wfamily{}$ satisfies TWC as in Definition~\ref{def:omegaMP},
WMI problems in class $\tree\WeightMI{\wfamily{}}$ are tractable, i.e., they can be solved in polynomial time.
\end{thm}
The proof to the above theorem is provided in the next two sections by construction where in Section~\ref{sec: MP-WMI} we proposed our WMI solver, called MP-WMI, operating on WMI problems in $\tree\WeightMI{\wfamily{}}$ with its complexity analysis in Section~\ref{sec: complexity}.
A summary of the WMI problem classes is shown in Figure~\ref{fig: wmi problem classes}.

 \section{Message-Passing WMI}
\label{sec: MP-WMI}

Message passing on tree-structured graphs has achieved 
remarkable attention
in the PGM literature~\cite{pearl1988probabilistic,kschischang2001factor}.
Its classical formulation and efficiency relies on compact factor representations allowing easy computations.
However, 
adapting existing message-passing algorithms  to WMI inference is non-trivial.
This is due to the fact that inference is computed in a hybrid structured space with logical and arithmetic constraints.
We present our message-passing scheme by first deriving a factorized representation of WMI problems.

\subsection{Factor Graph Representation of WMI}
\label{sec: factor graph representation of WMI}

In the literature of WMC, 
\textit{incidence graphs} are proposed to characterize the structure of problems defined by Boolean CNF formulas~\cite{samer2010algorithms}.
Incidence graphs are bipartite graphs with clause nodes and variable nodes,
where a clause and a variable node are joined by an edge if the variable occurs in the clause.
We derive the analogous representation for the more general SMT formulas, 
which we then turn into a factor graph of WMI problems. 

\begin{figure}[tb]
\centering
\begin{subfigure}[t]{0.15\textwidth}
\centering
\begin{tikzpicture}[grow=left, level/.style={sibling distance=12mm/#1, line width=\midlinewidth},level distance=11mm]
\scalebox{0.9}{
\node[circle, draw, line width=\midlinewidth](pt){$X_\team$}
    child {node [rectangle, draw, fill=cgreen](ft1){}
        child {node [circle, draw](x1){$X_1$}}
    }
    child {node [rectangle, draw, fill=cblue](ft2){}
        child {node [circle, draw](x1){$X_2$}}
    }
    child {node [rectangle, draw, fill=cred](ftb){}
        child {node [circle, draw](x1){$Z_B$}}
    };
\draw (ft1) edge[dashed, bend left, line width=\midlinewidth, ->] (pt);
\draw (ft2) edge[dashed, bend left, line width=\midlinewidth, ->] (pt);
\draw (ftb) edge[dashed, bend right, line width=\midlinewidth, ->] (pt);
}
\end{tikzpicture}
\end{subfigure}
\begin{subfigure}[t]{0.3\textwidth}
\centering
\scalebox{0.6}{
\begin{tikzpicture}
\draw[color=black, dashed, line width=\midlinewidth] ({\xscale*1}, {\yscale*(49/15)}) -- ({\xscale*1}, {0});
\draw[color=black, dashed, line width=\midlinewidth] ({\xscale*6}, {\yscale*(109/15)}) -- ({\xscale*6}, {0});
\draw[color=black, dashed, line width=\midlinewidth] ({\xscale*7}, {\yscale*(169/30)}) -- ({\xscale*7}, {0});
\draw[color=black, dashed, line width=\midlinewidth] ({\xscale*1}, {\yscale*(49/15) - \figoffset}) -- ({\xscale*1}, {0 - \figoffset});
\draw[color=black, dashed, line width=\midlinewidth] ({\xscale*6}, {\yscale*(109/15) - \figoffset}) -- ({\xscale*6}, {0 - \figoffset});
\draw[color=black, dashed, line width=\midlinewidth] ({\xscale*7}, {\yscale*(169/30) - \figoffset}) -- ({\xscale*7}, {0 - \figoffset});
\draw[color=black, line width=\midlinewidth] ({0}, {\yscale*1 - 2*\figoffset}) -- ({2*\xscale}, {\yscale*1 - 2*\figoffset});
\draw[color=black, line width=\midlinewidth] ({2*\xscale}, {\yscale*2 - 2*\figoffset}) -- ({7*\xscale}, {\yscale*2 - 2*\figoffset});
\draw[color=black, dashed, line width=\midlinewidth] ({2*\xscale}, {\yscale*2 - 2*\figoffset}) -- ({2*\xscale}, {0 - 2*\figoffset});
\draw[color=black, dashed, line width=\midlinewidth] ({7*\xscale}, {\yscale*2 - 2*\figoffset}) -- ({7*\xscale}, {0 - 2*\figoffset});
\draw[color=black, line width=\midlinewidth] plot [domain=0:1] ({\xscale*(\x)}, {\yscale*((7/30)*(\x)^3 - (7/5)*(\x)^2 + (7/5)*(\x) + (91/30))});
\draw[color=black, line width=\midlinewidth] plot [domain=1:6] ({\xscale*(\x)}, {\yscale*((4/5)*(\x)^2 - (24/5)*(\x) + (109/15))});
\draw[color=black, line width=\midlinewidth] plot [domain=6:7] ({\xscale*(\x)}, {\yscale*((-7/30)*(\x)^3 + (29/10)*(\x)^2 - (97/10)*(\x) + (172/15))});
\draw[color=black, line width=\midlinewidth] plot [domain=0:1] ({\xscale*(\x)}, {\yscale*((7/30)*(\x)^3 - (7/5)*(\x)^2 + (7/5)*(\x) + (91/30)) - \figoffset});
\draw[color=black, line width=\midlinewidth] plot [domain=1:6] ({\xscale*(\x)}, {\yscale*((4/5)*(\x)^2 - (24/5)*(\x) + (109/15)) - \figoffset});
\draw[color=black, line width=\midlinewidth] plot [domain=6:7] ({\xscale*(\x)}, {\yscale*((-7/30)*(\x)^3 + (29/10)*(\x)^2 - (97/10)*(\x) + (172/15)) - \figoffset});

\draw[-latex, line width=\incmidlinewidth] ( $(0,0) +(0pt,-\axisoffset)$ ) -- ( $(0,\ymax) +(0pt,\axisoffset)$ ) node[below right]{ \LARGE{$\msg{\sqbox{cgreen}}{X_{\mathcal{T}}}{}$} };

\draw[-latex, line width=\incmidlinewidth] ( $(0,0) +(-\axisoffset,0pt)$ ) -- ( $({13*(2/5)},0) +(\axisoffset,0pt)$ ) node[above]{\LARGE{$X_{\mathcal{T}}$} };

\draw[-latex, line width=\incmidlinewidth] ( $(0,-\figoffset) +(0pt,-\axisoffset)$ ) -- ( $(0,{\ymax-\figoffset}) +(0pt,\axisoffset)$ ) node[below right]{ \LARGE{$\msg{\sqbox{cblue}}{X_{\mathcal{T}}}{}$} };

\draw[-latex, line width=\incmidlinewidth] ( $(0,-\figoffset) +(-\axisoffset,0pt)$ ) -- ( $({13*(2/5)},-\figoffset) +(\axisoffset,0pt)$ ) node[above]{ \LARGE{$X_{\mathcal{T}}$} };

\draw[-latex, line width=\incmidlinewidth] ( $(0,{-2*\figoffset}) +(0pt,-\axisoffset)$ ) -- ( $(0,{\ymax-2*\figoffset}) +(0pt,\axisoffset)$ ) node[below right]{ \LARGE{$\msg{\sqbox{cred}}{X_{\mathcal{T}}}{}$} };

\draw[-latex, line width=\incmidlinewidth] ( $(0,{-2*\figoffset}) +(-\axisoffset,0pt)$ ) -- ( $({13*(2/5)},{-2*\figoffset}) +(\axisoffset,0pt)$ ) node[above]{ \LARGE{$X_{\mathcal{T}}$} };
\end{tikzpicture}
}
\end{subfigure}
\caption{\textbf{Factor graph}~(left) of formula $\gamma$ with two players and \textbf{piecewise polynomial messages}~(right) sent from the three factor nodes to variable node $X_\team$
when solving the WMI in Example~\ref{exa: WMI formulation of a skill matching system} by MP-WMI.}
\label{fig: team performance messages}
\end{figure}
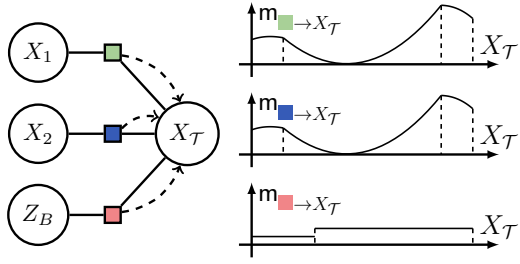

Recall that for the joint distribution represented by a WMI problem, 
the support is defined by the logical constraints and the unnormalized density is defined by weight functions.
In the following, we first factorize the SMT formula $\formula$ of a WMI problem $\WMI(\formula, \weights)$ in the class $\tree\WMI$:
\begin{align}
\label{eq: factorize formula}
    \formula = \bigwedge_{i \in \Va} \formula_i \land \bigwedge_{i,j \in \E} \formula_{i j}
\end{align}
where the set $\Va$ is the index set of variables and the set $\E$ is the index pairs of variables in the same clause.
Then a WMI problem can be conveniently represented as a bipartite graph, known as factor graph, 
that includes two sets of nodes: variable nodes $X_i$, and factor nodes $\factor{\clique}$, where $\clique$ denotes a factor scope, i.e.,  the set of indices of the variables appearing in it. 
A variable node $X_i$ 
is connected to a factor node $\factor{\clique}$ if and only if $i \in \clique$.
Specifically, the factors are defined as follows:
\begin{shrinkeq}{-0ex}
\begin{align}
\label{eq: definition of factors}
    \factor{\clique}(\x_\clique)
    = \!\!\!\prod_{\Gamma \in \Clause(\formula_\clique)} \!\!\!\id{\x_\clique \models \Gamma}
    \!\!\!\prod_{\ell \in \Literal(\Gamma)} \!\!\!w_\ell(\x_\clique)^{ \id{\x_\clique \models \ell} }
\end{align}
\end{shrinkeq}
where $\x_\clique$ denotes the restriction of $\x$ to the variables in factor $\factor{\clique}$ and analogously $\formula_\clique$ is the restriction of formula $\formula$ to the clauses over the variables in $\clique$.
Here, the set of clauses in the SMT formula $\formula$ is denoted by $\Clause(\formula)$,
and the set of literals in a clause $\Gamma$ is denoted by $\Literal(\Gamma)$.
Intuitively,
the factors include the parameterized densities as in the classic PGM literature, here represented by the per-literal weights,
but also the structure enforced by the logical constraints in the SMT formula, via the indicator functions.
Figure~\ref{fig: team performance messages} shown an example of a factor graph.

As in every tree-shaped factor graphs, we define an unnormalized joint distribution corresponding to the WMI problem in the form of a product of factors as follows.
\begin{align}
\label{eq: factorized joint distribuiton}
    p(\x)
    = \prod_\clique \factor{\clique} (\x_\clique)
    = \prod_{i \in \Va} \factor{i}(X_i) \prod_{i, j \in \E} \factor{ij}(X_i, X_j)
\end{align}
By construction, it is easy to see that the normalization constant of such a distribution equals computing the corresponding weighted model integral.
\begin{pro}
\label{prop: partition function equals to WMI}
Given a problem $\WMI(\formula, \weights)$ in $\tree\WMI$,
let $p(\x)$ being the unnormalized joint distribution as defined in Equation~\ref{eq: factorized joint distribuiton}.
Then the partition function of distribution $p(\x)$ is equal to $\WMI(\formula, \weights)$.
\end{pro}

\subsection{Message-Passing Scheme}

Deriving a message-passing scheme for WMI poses unique and considerable challenges.
First, different from discrete domains, on continuous or hybrid domains one generally does not have universal and compact representations for messages, and logical constraints in WMI make it even harder to derive such representations.
Moreover, marginalization over real variables requires integration over polytopes, which
is already \#P-hard~\cite{dyer1988complexity}.
The integration poses the problem of whether the messages defined are integrable and how hard it is to perform the integration.
In the following part, we will present our solutions to these challenges by first describing a general message-passing scheme for WMI and then investigating of which form the messages are, given certain weight families.

Given the factorized representation of WMI in Section~\ref{sec: factor graph representation of WMI}, our message-passing scheme, 
called MP-WMI and summarized in Algorithm~\ref{alg: MP-WMI weighted message passing}, 
comprises exchanging messages between nodes in the factor graph.
Before the message passing starts, we choose an arbitrary node in the factor graph as root and orient all edges away from the root to define the message sending order.
MP-WMI operates in two phases: an upward pass and a downward one. 
First, we send messages up from the leaves to the root (upward pass) such that each node has all information from its children and then we incorporate messages from the root down to the leaves (downward pass) such that each node also has information from its parent.
The messages are formulated as follows.
\begin{pro}
\label{pro: recursive messages}
Both messages $\msg{\factor{ij}}{X_i}{}$ from factor node to variable node 
and messages $\msg{X_i}{\factor{ij}}{}$ from variable node to factor node
have iterative formulations as follows.
\begin{enumerate}[label=(\roman*),noitemsep,topsep=0pt]
    \item $\msg{\factor{ij}}{X_i}{}(x_i) = \int \factor{ij}(x_i, x_j) \cdot \msg{X_j}{\factor{ij}}{}(x_j) ~ d x_j$;
    \item $\msg{X_i}{\factor{\clique}}{}(x_i) = \prod_{\factor{\clique^\prime} \in \neigh(X_i) \backslash \factor{\clique}} \msg{\factor{\clique^\prime}}{X_i}{}(x_i)$.
\end{enumerate}
\end{pro}
For the start of sending messages, when a leaf node is a variable node $X_i$, the message that it sends along its one and only edge to a factor $\factor{\clique}$ is $\msg{X_i}{\factor{\clique}}{}(c_i) = 1$; 
in the case when a leaf node is a factor node $\factor{i}$, the message from the factor node $\factor{i}$ to a variable node $X_i$ is $\msg{\factor{i}}{X_i}{}(x_i) = \factor{i}(x_i)$.
Even though the weight function family is not specified here, it can be shown that when the integration in Proposition~\ref{pro: recursive messages} is well-defined, i.e., the integrands are integrable, then the messages are univariate piecewise functions, which is a striking difference with classical message-passing schemes.
\begin{pro}
\label{prop: piecewise function messages}
For any problem in $\tree\WMI$,
the messages as in Proposition~\ref{pro: recursive messages}
are univariate piecewise functions. 
\label{pro: piecewise polynomial}
\end{pro}

\begin{algorithm}[tp]
\caption{\textbf{MP-WMI}($\theory$)
}
\label{alg: MP-WMI weighted message passing}
\begin{algorithmic}[1]
\STATE $\V_{\mathsf{up}} \leftarrow$ sort variable nodes in factor graph
\FOR[upward pass]{\textbf{each} $X_{i}\in\V_{\mathsf{up}}$} 
\STATE \textsf{send-message}($X_i$,$\factor{i, \pa(i)}$) 
\STATE \textsf{send-message}($\factor{i, \pa(i)}$, $X_{\pa(i)}$)
\ENDFOR
\STATE $\V_{\mathsf{down}} \leftarrow$ 
sort nodes in set $\V_{\mathsf{up}}$ in reverse order
\FOR[downward pass]{\textbf{each} $X_{i}\in\V_{\mathsf{down}}$}
\FOR{\textbf{each} $X_{c}\in\ch(X_{i})$}
\STATE \textsf{send-message}($X_i$, $\factor{ic}$)
\STATE \textsf{send-message}($\factor{ic}$, $X_c$)
\ENDFOR
\ENDFOR
\STATE \textbf{return}
$\{ \msg{X_i}{\factor{s}}{}, \msg{\factor{s}}{X_i}{} \}_{(x_i, \factor{s}) \in \E}$
\end{algorithmic}
\end{algorithm}

The specific form of messages also depends on the chosen weight function family as mentioned in Section~\ref{sec: tractable}. For example, when the weight functions are chosen to be polynomials, the messages are piecewise polynomials, as in the example in Figure~\ref{fig: team performance messages}. 
We show how to compute the piecewise polynomial messages in Algorithm~\ref{alg: send messages} with functions \textit{critical-points} and \textit{get-msg-pieces} as subroutines to compute the numeric and symbolic integration bounds for the message pieces. 
Both of them can be efficiently implemented, see \citet{zeng2019efficient} for details.
The actual integration of the polynomial pieces can be efficiently performed symbolically, as supported by many scientific computing packages.

When MP-WMI terminates, the information stored in the obtained messages is sufficient to compute the unnormalized marginals for each variable and it is independent of the choice of root.
Moreover, the integration of unnormalized marginals equals to $\WMI(\formula, \weights)$.

\begin{pro}
\label{prop: messages and unnormalized distribution and WMI}
Let $\formula$ be an SMT formula with a tree factor graph and with per-literal weights $\weights$.
For any variable $X_i$, the unnormalized marginal $p(x_i)$ is
\begin{align*}
    p(x_i) = \prod\nolimits_{\factor{\clique} \in \neigh(X_i)} \msg{\factor{\clique}}{X_i}{}(x_i).   
\end{align*}
Moreover, the partition function for any $x_i$ is the WMI of SMT formula $\formula$, i.e., $\WMI(\formula, \weights) = \int_{x_i} p(x_i) d x_i$.
\end{pro}

\begin{algorithm}[tp]
\caption{\textbf{\textsf{send-message}}($s$, $t$)}
\label{alg: send messages}
\begin{algorithmic}[1]
\IF{$s = X_i$ and $t = \factor{ij}$}
\STATE \textbf{Return}
$\prod_{\factor{s^\prime} \in \neigh(X_i) \backslash \factor{ij}} \msg{\factor{s^\prime}}{X_i}{}$
\ELSIF{$s = \factor{ij}$ and $t = X_i$}
\STATE $\mathcal{P} \leftarrow$ \textsf{critical-points}($\msg{X_j}{\factor{ij}}{}, \formula_{ij}$)
\STATE $\mathcal{I} \leftarrow$ \textsf{intervals-from-points}($\mathcal{P}$)
\FOR{interval $I \in \mathcal{I}$ consistent with formula $\formula_{ij}$}
\STATE $ \langle l_s, u_s, p \rangle \leftarrow$
\textsf{get-msg-pieces}($\msg{X_j}{\factor{ij}}{}, I, w$)
\STATE $p^\prime(x_i) \leftarrow \int_{l_{s}}^{u_{s}} p(x_i, x_j) ~d x_j$
\STATE $\msg{\factor{ij}}{X_i}{} \leftarrow \msg{\factor{ij}}{X_i}{} + \id{x_i \in I} \cdot p^\prime(x_i) $
\ENDFOR
\ENDIF
\STATE \textbf{return} $\msg{s}{t}{}$
\end{algorithmic}
\end{algorithm}

\subsection{Amortization}
\label{sec: Amortization}

We will show that by leveraging the messages pre-computed in MP-WMI, we are able to speed up (amortize) inference time over multiple queries on formula $\formula$.
More specifically, when answering queries that do not change the tree structure in the factor graph of formula $\formula$,
we only need to update messages that are related to the queries while other messages are pre-computed.
Some examples are SMT queries on a node variable or queries over a pair of variables that are connected by an edge in the factor graph,
since these queries either add leaf nodes or do not change existing nodes.
Thus we can reuse the local information encoded in messages. 

\begin{pro}
Let $\WMI(\formula, \weights)$ be a problem in $\tree\WMI$,
and $\Phi$ be an SMT query over a factor $\factor{s}^*$ involving a variable $X_{i}\in\X$.
Given pre-computed messages $\{ \msg{\factor{\clique}}{X_i}{}\}_{\factor{\clique} \in \neigh(X_i)}$,
\begin{align*}
    \WMI(\formula\wedge\Phi)=\int_{\mathbb{R}} & \msg{\factor{s}^*}{X_i}{*}(x_i) \cdot \\
    & \prod_{\factor{s} \in \neigh(X_i) \backslash \factor{s}^*} \msg{\factor{s}}{X_i}{}(x_i)  d x_{i}
\end{align*}
with message $\msg{\factor{s}^*}{X_i}{*}$ computed over factor $\factor{s}^*(\x_s) := \factor{s}(\x_s) \cdot \id{\x_s \models \Phi}$ as in Proposition~\ref{pro: recursive messages}.
\label{prop: wmi-msg-any}
\end{pro}
Pre-computing messages can dramatically speed up inference by amortization, as we will show in our experiments, especially when traversing the factor graph is expensive or the number of queries is large.

\section{Complexity Analysis}
\label{sec: complexity}

This section provides a complexity analysis of our proposed WMI solver MP-WMI.
Given the SMT formula $\formula$ with a tree factor graph with a chosen root node,
each factor node would be traversed exactly once in each phase of the message-passing scheme.
We denote the set of directed factor nodes by $\F := \{\overrightarrow{\factor{s}} \} = \{ \factor{s}^{\up}, \factor{s}^{\down} \mid \factor{s} \in \Va \}$
where $\factor{s}^{\up}$ denotes the factor node $\factor{s}$ visited in the upward pass and $\factor{s}^\down$ denotes the one visited in downward pass respectively.

To characterize the message-passing scheme, we define a nilpotent matrix $A$ as follows.
The matrix $A \in \mathbb{N}^{|\F| \times |\F|}$ has both its columns and rows denoted by the factor nodes in set~$\F$.
At each column denoted by $\overrightarrow{\factor{s}}$, only entries at rows denoted by factor nodes visited right after $\overrightarrow{\factor{s}}$ are non-zero.

\begin{pro}
\label{prop: nilpotent matrix order}
The nilpotent matrix $A$ as described above has its order at most the diameter of the factor graph. 
\end{pro}

Next we show how to define the non-zero entries in matrix $A$ with parameters about the SMT formulas in WMI problems.
\begin{pro}
\label{prop: upper bound of msg pieces}
Suppose that the two variables $X_i$ and $X_j$ are connected in the factor graph by a factor $\factor{ij}$ associated with a sub-formula $\formula_{ij}$ of size $c$, then in MP-WMI:
\begin{enumerate}[label=(\roman*),noitemsep,topsep=0pt]
    \item the number of pieces in message $\msg{X_i}{\factor{ij}}{}$ is bounded by $\sum m_s$, where $m_s$ is the number of pieces in message $\msg{\factor{s}}{X_i}{}$ with $\factor{s} \in \neigh(X_i) \backslash \factor{ij}$;
    \item the number of pieces in message $\msg{\factor{ij}}{X_j}{}$ is bounded by $2mc + c^2$ with $m$ being the number of pieces in message $\msg{X_i}{\factor{ij}}{}$.
\end{enumerate}
\end{pro}

Now we show how to use the matrix $A$ to bound the number of pieces in messages.
We define the non-zero entries in the nilpotent matrix $A$ to be $2c$ with $c$ being a constant that bounds the size of sub-formulas associated to factors.
Define a vector $v^{(t)} \in \mathbb{N}^{|\overrightarrow{\E}|}$ 
for the state of the message-passing scheme at step $t$ -- by state it means that each entry in vector $v^{(t)}$ is denoted by a factor node in set $\F$ and the entry denoted by $\overrightarrow{\factor{s}}$ bounds the number of pieces in the message sent to $\factor{s}$ 
in the MP-WMI.
For the initial state vector $v^{(0)}$, it has all non-zero entries to be $c$, the constant bounding the sub-formula size, and these entries are those denoted by $\overrightarrow{\factor{s}} = \factor{s}^+$ with factor node $\factor{s}$ connected with a leaf.
\begin{pro}
\label{prop: recursive upper bound for msg pieces}
Let $A$ be the nilpotent matrix and $v$ the initial state vector as described above. Also let $v^{(t)} := A v^{(t-1)} + c^2 \cdot \sgn(A v^{(t-1)})$ with $\sgn$ being the sign function.
Then each entry in vector $v^{(t)}$ denoted by $\overrightarrow{\factor{s}}$ bounds the number of pieces in the message $\msg{X_i}{\factor{s}}{}$ received by factor $\factor{s}$ from some variable node $X_i$ at step $t$ in 
MP-WMI.
\end{pro}

\begin{pro}
\label{prop: bounds for number of message pieces}
Let $A$ be the nilpotent matrix and $v^{(t)}$ the state vectors as described above.
The total number of pieces in the all the messages is bounded by $\parallel \sum_{t=0}^{d} v^{(t)} \parallel_1$ with $d$ being the diameter of the factor graph.
Moreover, it holds that $\parallel \sum_{t = 0}^{d} v^{(t)} \parallel_1$ is of $\bigO((4 n c)^{2d + 2})$.
\end{pro}
This gives the worst-case total number of message pieces in MP-WMI.
From Proposition~\ref{prop: bounds for number of message pieces}, it holds that the problems in class $\tree\WeightMI{\wfamily{}}$ with the weight function family $\wfamily{}$ satisfying TWC are tractable to MP-WMI,
since the complexity of MP-WMI is the total number of message pieces multiplied by the symbolic integration cost of each piece,
which is tractable for functions in family $\wfamily{}$ by definition.
This finishes the constructive proof for Theorem~\ref{thm:omegaMI-in-omegaMP} in Section~\ref{sec: tractable}.
Notice the complexity of WMI problems depends on the graph structures.
In our experiments, we will compare solvers on WMI problems with three representative problem classes with different factor graph diameters.

\section{Related Work}
\label{sec: related work}

WMI generalizes weighted model counting (WMC)~\cite{sang2005performing} to hybrid domains~\cite{belle2015probabilistic}.
WMC is one of the state-of-the-art approaches for inference in many discrete probabilistic models. 
Existing exact WMI solvers for arbitrarily structured problems include DPLL-based search with numerical~\cite{belle2015probabilistic,morettin2017efficient,morettin2019advanced} or symbolic integration~\cite{braz2016probabilistic} and compilation-based algorithms~\cite{kolb2018efficient,zuidberg2019exact}.

Motivated by their success in WMC, \citet{belle2016component} present a caching scheme for WMI that allows reusing computations at the cost of not supporting algebraic constraints between variables.
Different from usual, \citet{merrell2017weighted} adopt Gaussian distributions,  while~\citet{zuidberg2019exact} fixed univariate parametric assumptions for weight functions.
Closest to our MP-WMI, SMI~\cite{zeng2019efficient} is an exact solver which leverages context-specific independence to perform efficient search and operates on tree-shaped primal graphs.
Many recent efforts in WMI converged in the \textit{pywmi}~\cite{kolb2019pywmi} python framework.

\begin{figure*}[!t]
\centering
\includegraphics[width=.32\textwidth]{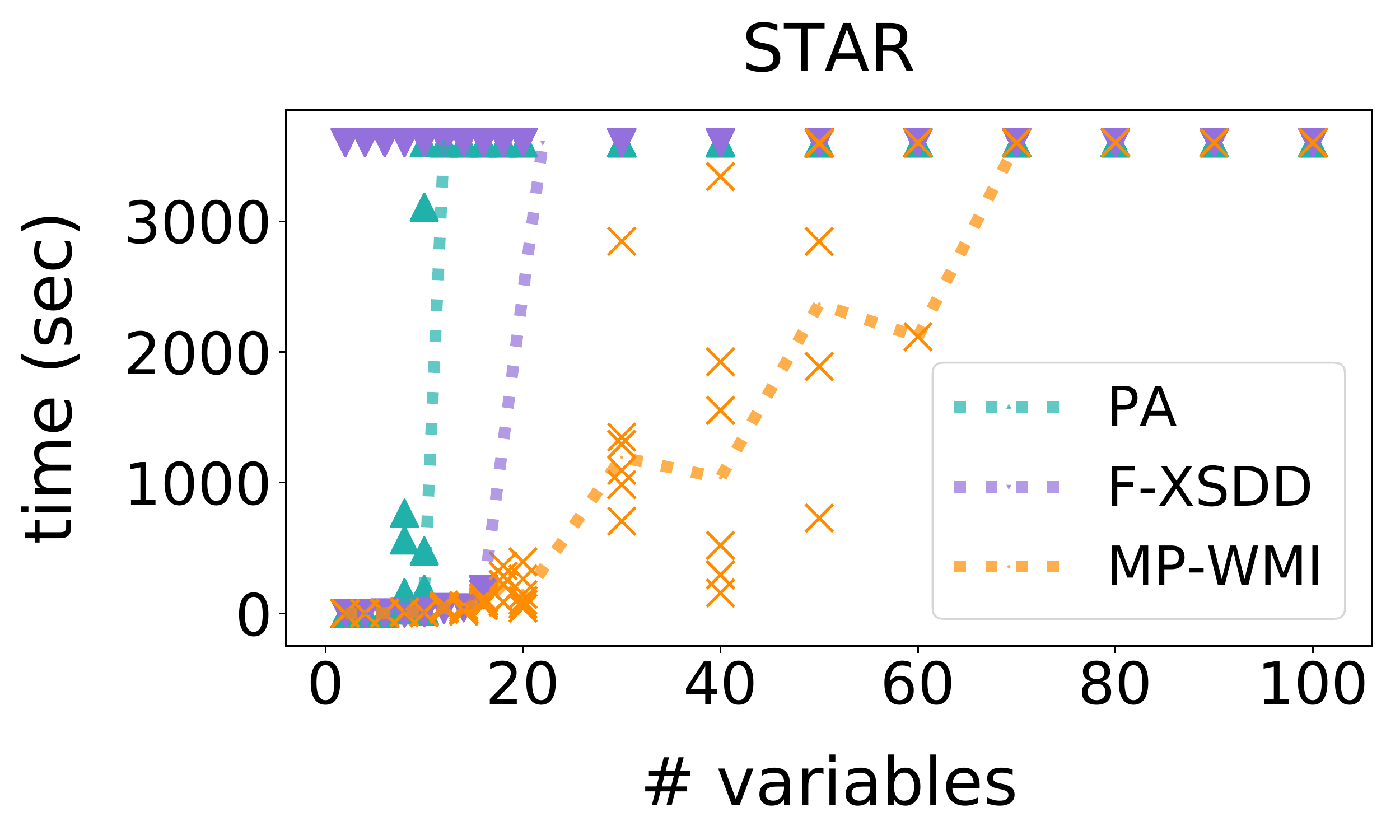}\hfill
\includegraphics[width=.32\textwidth]{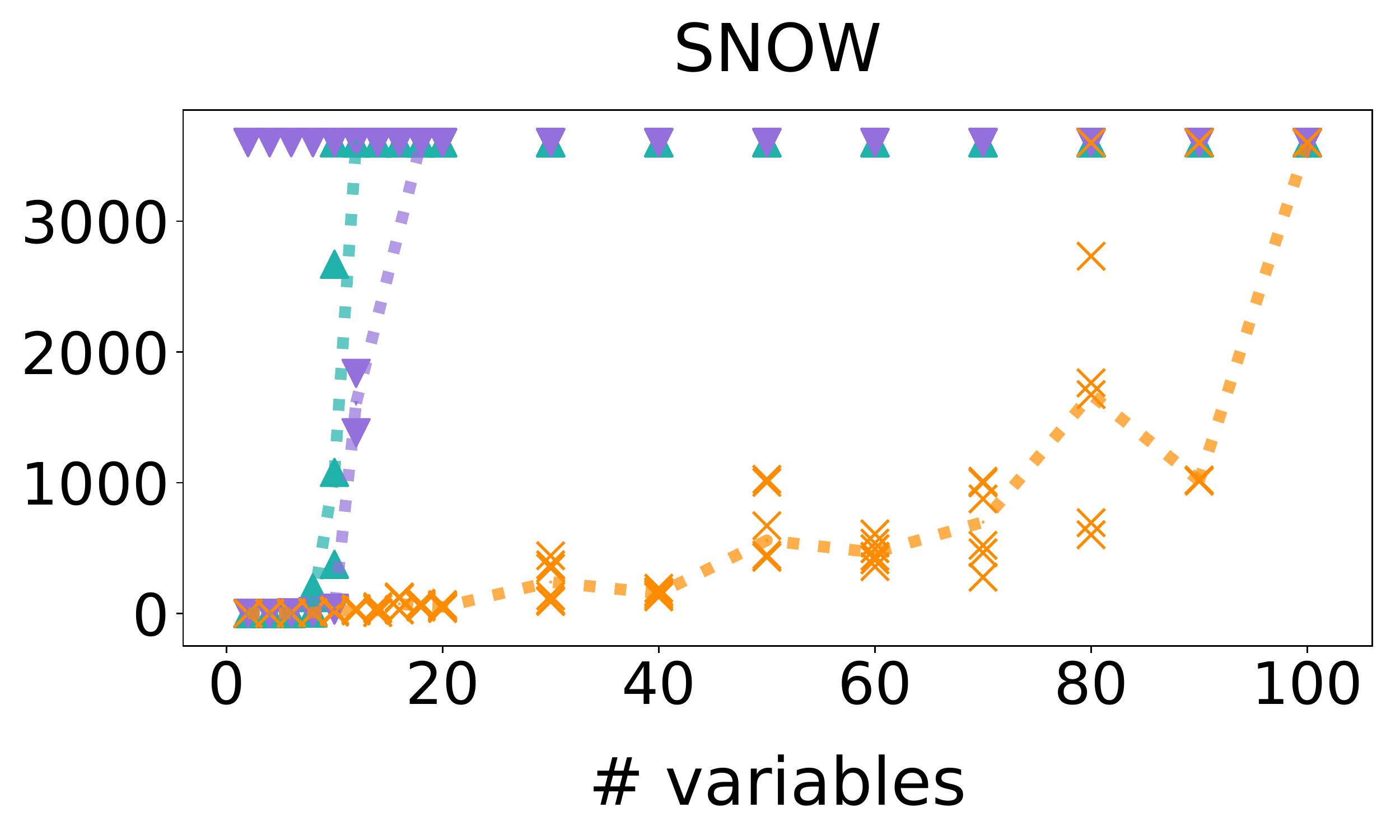}\hfill
\includegraphics[width=.32\textwidth]{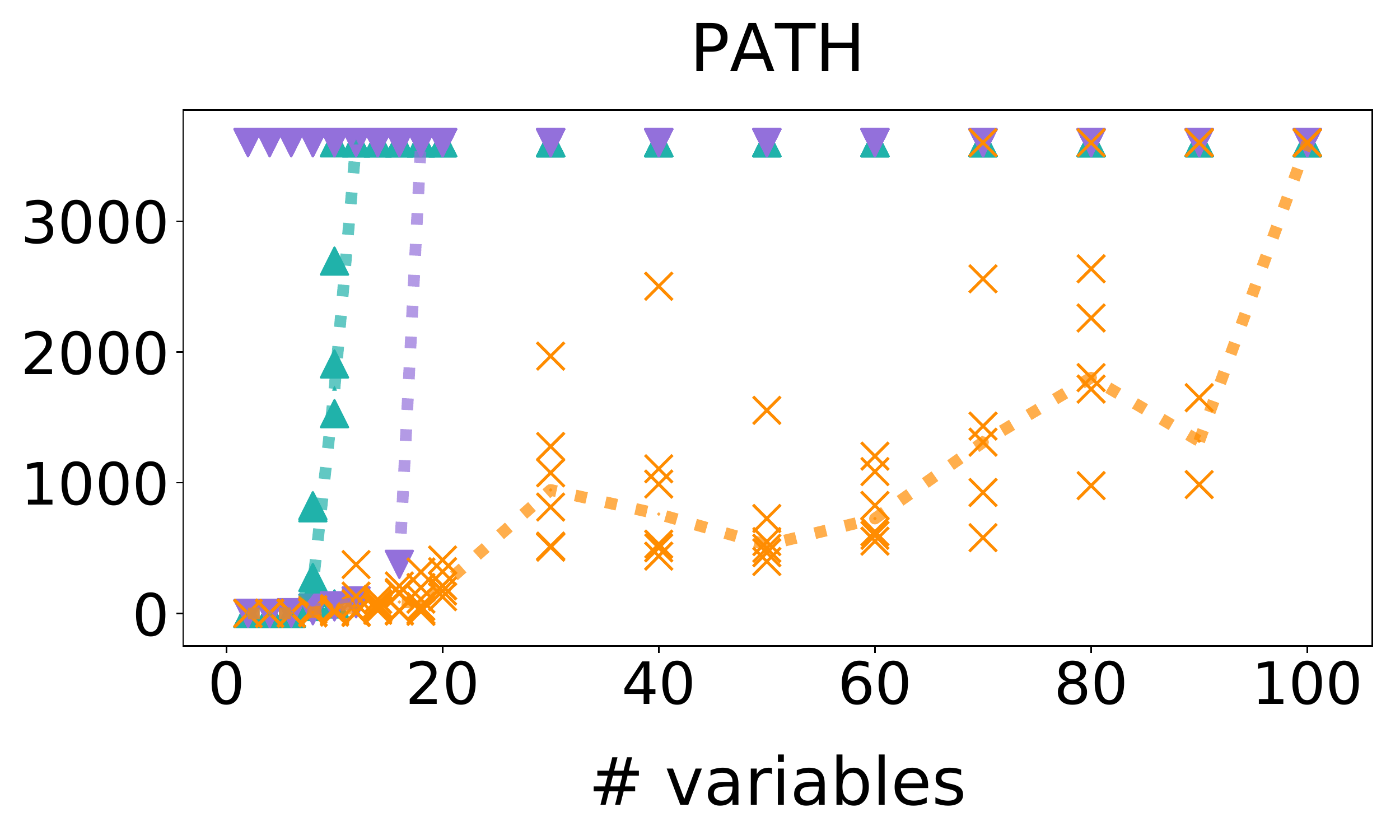}
\caption{Results of the comparison between MP-WMI, WMI-PA and F-XSDD on WMI problems with tree dependencies. 
In this setting, MP-WMI remarkably scales to problems having up to $60$ variables on STAR, while solving SNOW and PATH problems having up to $90$ variables, considerably ``raising the bar'' for the size of tractable WMI inference problems.
}
\label{fig:results}
\end{figure*}

Tree-shaped dependency structures, as the ones characterizing our $\tree\WeightMI{\boldsymbol\Omega}$ class, naturally arise in many fields, such genetics~\cite{nei2000molecular}, system analysis~\cite{vesely1981fault}, linguistics~\cite{petrov2006learning}, and telecommunications~\cite{leon2003communication}.
Moreover, thanks to their appealing mathematical properties, trees serve as practical approximations of non tree-shaped  problems~\cite{rubinstein1983signal,robins2000improved,binev2004fast}.

\begin{figure*}[!t]
    \centering
    \includegraphics[width=.99\textwidth]{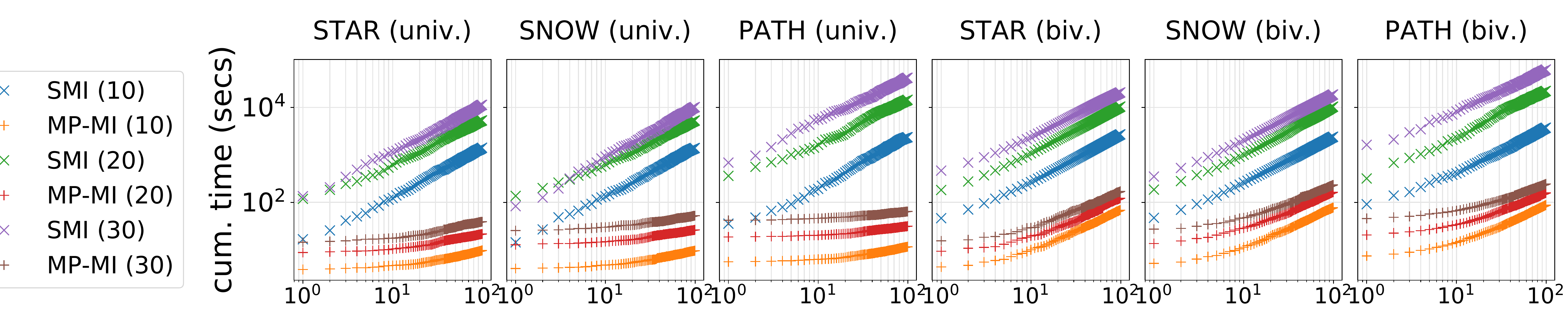}
    \caption{Log-log plot of cumulative time (seconds, y-axis) for MP-WMI (orange, red, brown) and SMI (blue, green, purple) over STAR, SNOW and PATH primal graphs (see text) with 10, 20 and 30 variables for increasing numbers of univariate and bivariate queries (x-axis). 
    For every class, MP-WMI takes up to two order of magnitude less time when amortizing $100$ queries, while being faster than SMI on a single query.
    }
    \label{fig:amortize}
\end{figure*}

Message-passing schemes have been widely used for developing exact and approximate inference algorithms for probabilistic graphical models on discrete~\cite{kschischang2001factor}, continuous~\cite{guo2019marginal,wang2017stein} and hybrid domains~\cite{gogate2012approximate}.
Our amortization scheme is closely related to the reuse of local computation in the join tree algorithm~\cite{huang1996inference,lepar2013comparison}, which has never been explored in hybrid domains for WMI inference, however.
Similarly to us, \citet{gamarnik2012belief} adopts piecewise polynomial messages, specifically piecewise-linear convex functions, in a belief propagation scheme for non-probabilistic min-cost network flow problems.

Research on learning WMI distributions from data is at its early stages.
Parameter learning for piecewise constant densities has been addressed in~\cite{belle2015probabilistic}. Recently, an approach for jointly learning the structure and parameters of a WMI problem has been proposed in~\cite{morettin2020learning}. Developing faster inference algorithms is thus beneficial in learning scenarios as, typically, learning a full model requires numerous calls to an inference procedure. 
WMI inference is closely related to probabilistic program inference, where complex arithmetic and logical constraints are induced by the program structure or its abstraction~\citep{HoltzenUAI17,HoltzenICML18}.

\section{Experiments}
\label{sec:exp}

In this Section, we aim to answer the following research questions:\footnote{Our implementation of MP-WMI  and the code for reproducing our empirical evaluation can be found at \url{https://github.com/UCLA-StarAI/mpwmi}.}
{\bf Q1)} Can we effectively scale WMI inference with MP-WMI?
{\bf Q2)} How beneficial is inter-query amortization with MP-WMI?

To answer~{\bf Q1}, we generated a benchmark of WMI problems with tree-shaped primal graphs of different diameters:
star-shaped graphs~(STAR),
complete ternary trees~(SNOW)
and linear chains~(PATH).
These structures were originally investigated by the authors of SMI and are prototypical of 
tree shapes that can be encountered in real-world scenarios such as 
phylogenetic trees~\cite{nei2000molecular}, hierarchies in file and networks systems~\cite{vesely1981fault}, and natural language grammars~\cite{petrov2006learning}.

We sampled random SMT formulas with $N$ variables with the tree structures described above
and polynomial weights
mapping a subset of
literals to a random non-negative polynomials.
We generated problems with $N$ ranging  
from $2$ to $20$ with step size $2$, and from $20$ to $100$ with step size $10$.
We compared our MP-WMI python implementation against the following baselines: WMI-PA~\cite{morettin2019advanced}, a solid general-purpose WMI solver exploiting SMT-based predicate abstraction techniques that is less sensitive to the problem structure; and F-XSDD(BR)~\cite{zuidberg2019exploit}, a compilation-based solver achieving state-of-the-art results in several WMI benchmarks.

Fig.~\ref{fig:results} shows that,
with timeout being an hour,
our proposed solver MP-WMI is able to scale up to 60 variables for STAR problems and up to 90 variables for SNOW and PATH problems, while the other two solvers stop at problem size 20 for all three classes. 
Note that the results are in line with those reported in~\cite{zuidberg2019exploit}.
This answers {\bf Q1} affirmatively, raising the bar of the size of WMI problems that can be solved exactly up to 100 variables.

We tackle {\bf Q2} by comparing MP-WMI with SMI~\cite{zeng2019efficient} on tree-structured MI problems.
SMI is a search-based MI solver that has been shown to be efficient for such problems. 
WMI-PA, F-XSDD and the SGDPLL(T)~\cite{braz2016probabilistic} solver are not included in the comparison since they were already shown in \citet{zeng2019efficient} to not be competitive on such problems.
The synthetic SMT formulas range over $n\in\{10, 20, 30\}$ variables with tree factor graphs being STAR, SNOW and PATH.
We generate $100$ univariate or bivariate random queries for each MI problem.

Figure~\ref{fig:amortize} shows the cumulative runtime of answering random queries by both solvers. 
As expected, MP-WMI takes a fraction of the time of SMI  (up to two order of magnitudes) to answer 100 univariate or bivariate queries in all experimental scenarios, since it is able to amortize inference \textit{inter-query}.
More surprisingly, by looking at the first point of each curve, we can tell that MP-WMI is even faster than SMI to compute a single query.
This is because SMI solves polynomial integration numerically, by reconstructing the univariate polynomials before the numeric integration via interpolation, e.g., Lagrange interpolation; while in MP-WMI we adopt symbolic integration. 
Hence the complexity of the former is always quadratic in the degree of the polynomial, while for the latter the average case is linear in the number of monomials in the polynomial to integrate, which in practice might be much less than the degree of the polynomial.

\section{Conclusions}
\label{sec:conclusions}

In this paper, we theoretically traced the boundaries of tractable WMI inferece and proposed a novel exact WMI solver based on message-passing, MP-WMI,
which
is efficient on a rich class of tractable WMI problems with tree-shaped factor graphs, the largest known so far.
Furthermore, MPWMI dramatically reduces the answering time of multiple queries by amortizing local computations and allows to compute all marginals and moments simultaneously.

We believe this provides a theoretical and algorithmic stepping stone needed to device principled approximate WMI inference schemes that can scale even further to larger and non tree-shaped problem structures.

\scalebox{.01}{non omnes arbusta iuvant humilesque myricae}\vspace{-5pt}
\section*{Acknowledgements}
This work is partially supported by NSF grants \#IIS-1943641, \#IIS-1633857, \#CCF-1837129, DARPA XAI grant \#N66001-17-2-4032, a Sloan fellowship, and gifts from Intel and Facebook Research.

The authors would like to thank Arthur Choi for insightful discussions about message passing routines and generalizations in the PGM literature.
\bibliography{ref}
\bibliographystyle{icml2020}

\clearpage
\appendix
\section{Proofs}

\subsection{THEOREM~\ref{thm: sharp p hard of 2-Clause formula}}
\begin{proof}(Theorem~\ref{thm: sharp p hard of 2-Clause formula})
The proof is done by reducing the \#P-complete problem \#2SAT over a 2SAT formula $\formula_{\Bool}$ to an MI problem on a 2-Clause \smtlra~formula $\formula$.

By the Boolean-to-real reduction from \cite{zeng2019efficient}, 
there exists an \smtlra~formula $\formula$ over real variables only such that $\MI(\formula_{\Bool}) = \MI(\formula)$.
The formula $\formula$ can be obtained in the following way.
Any Boolean literal $B$ or $\neg B$ in propositional formula $\formula_{\Bool}$ is substituted by $\lra$ literals $Z_B > 0$ and $Z_B < 0$ respectively where the real variable $Z_B$ is an auxiliary real variable with bounding box $(Z_B \ge -1) \land (Z_B \le 1)$.
Denote the formula after replacement by $\formula^\prime$. Then we have formula $\formula$ as follows.
\begin{equation*}
    \formula = \formula^\prime \land \bigwedge_{B \in \vars(\formula_{\Bool})} (Z_B \ge -1) \land (Z_B \le 1)
\end{equation*}
For each clause in formula $\formula$, since it contains at most two Boolean variables before substitution, it also contains at most two real variables now. 
Therefore formula $\formula$ is a 2-Clause \smtlra~formula over real variables only.
Moreover, the reduction guarantees that $\MI(\formula) = \MI(\formula_\Bool)$ where $\MI(\formula_\Bool)$ is the number of satisfying assignments to $\formula_\Bool$ by the definition of WMI.
Thus, computing MI of a 2-Clause \smtlra~ formula over real variables is \#P-hard.
\end{proof}

\subsection{THEOREM~\ref{thm:rest-rho}}
\begin{proof}(Theorem~\ref{thm:rest-rho})
When the weight function family $\wfamily{} = \wfamily{\mathsf{SMI}}$,
by the WMI-to-MI reduction process in~\citet{zeng2019efficient},
any WMI problem in $\tree\WeightMI{\wfamily{}}$ can be reduced to an MI problem in class $\tree\MI$.

We prove the other way by contradiction.
Suppose that there exists a WMI problem $\nu = \WMI(\formula, \weights) \in \tree\WeightMI{\wfamily{}}$ with a per-literal weight function $w_\ell \notin \wfamily{\mathsf{SMI}}$ such that $\rho(\nu) \in \tree\MI$.
Since the per-literal weight function $w_\ell \notin \wfamily{\mathsf{SMI}}$,
from the definition of $\wfamily{\mathsf{SMI}}$, it holds that $\ell$ is a bivariate literal defined in a clause $\Gamma$ which is a conjunction of more than one distinct literals, i.e., $\Gamma = \ell \vee \bigvee_{i=1}^k \ell_i, k \geq 1$ with $\ell \neq \ell_i, \forall i = 1, \cdots, k$.
During the reduction, a clause $\Gamma^\prime = \ell \Rightarrow \land_j^m \theta_j$ is conjoined to the formula $\formula$ to encode the weight function $w_\ell$ with at least one auxiliary variable in formula $\theta_j$.
Then there are at least three distinct variables in clause $\Gamma^\prime$ since given the form of clause $\Gamma$, clause $\Gamma^\prime$ can not be further simplified by resolution.
This causes a loop in the primal graph of the reduced MI problem $\rho(\nu)$, which contradicts the assumption that $\rho(\nu) \in \tree\MI$.
Therefore, if $\forall \nu \in \tree\WeightMI{\wfamily{}}$, $\rho(\nu) \in \tree\MI$, then $\wfamily{} \subseteq \wfamily{\mathsf{SMI}}$.
\end{proof}

\subsection{PROPOSITION~\ref{prop: partition function equals to WMI}}
\begin{proof}(Proposition~\ref{prop: partition function equals to WMI})
Recall that given a WMI problem with SMT formula $\formula$ over real variables only, the WMI can be computed as follows by the definition of WMI in Equation~\ref{eq:wmi}.
\begin{align*}
    \WMI(\formula, \weights) = \int_{\x \models \formula} \prod_{\ell \in \Literal(\formula)} w_\ell(\x)^{\id{\x \models \ell}} ~d \x
\end{align*}
Notice that this is equivalent to integrating on domain $\R^{\mid \X \mid}$ over the integrand
$f(\x) = \id{\x \models \formula} \prod_{\ell \in \Literal(\formula)} w_\ell(\x)^{\id{\x \models \ell}}$.
Next, we show how to factorize over the integrand $f(\x)$ based on the factorization on formula $\formula$ in Equation~\ref{eq: factorize formula}.
First, for the indicator function, we have that
\begin{align*}
    \id{\x \models \formula} = \prod_{\clique} \id{\x_\clique \models \formula_\clique} = \prod_{\clique} \prod_{\Gamma \in \Clause(\formula_\clique)} \id{\x_\clique \models \Gamma}.    
\end{align*}
Moreover, it holds that
\begin{align*}
    \prod_{\ell \in \Literal(\formula)} w_\ell(\x)^{\id{\x \models \ell}}
    = \prod_{\clique} \prod_{\Gamma \in \Clause(\formula)} \prod_{\ell \in \Literal(\Gamma)} w_{\ell}(\x_\clique)^{\id{\x_\clique \models \ell}}.
\end{align*}
Together they complete the proof that the integrand $f(x)$ here equals to the unnormalized joint distribution $p(\x)$ defined in Equation~\ref{eq: factorized joint distribuiton} and therefore the partition function of distribution $p(\x)$ equals to the WMI of formula $\formula$.
\end{proof}

\subsection{PROPOSITION~\ref{prop: piecewise function messages}}
\begin{proof}(Proposition~\ref{prop: piecewise function messages})
This follows by induction on the message-passing scheme.
Consider the base case of the messages sent by leaf nodes.
When the leaf node is a variable node $X_i$, by definition the messages it sends to a factor node $\factor{\clique}$ is $\msg{X_i}{\factor{\clique}}{}(X_i) = 1$;
when the leaf node is a factor node $\factor{i}$, by definition the messages it sends to the variable node $X_i$ is $\msg{\factor{i}}{X_i}{}(X_i) = \factor{i}(X_i)$.
By the definition of factor functions in Equation~\ref{eq: definition of factors}, the function $\factor{i}$ is a univariate piecewise function in variable $X_i$ with pieces defined by the logical constraints in formula $\formula_i$ as in Equation~\ref{eq: factorize formula}.
Then it holds that messages sent from the leaf nodes in the message-passing scheme are piecewise function.

Further, by the recursive formulation of messages in Proposition~\ref{pro: recursive messages},
since the piecewise functions are close under product, messages sent from variable nodes to factor nodes are again univariate piecewise functions;
for messages $\msg{\factor{\clique}}{X_i}{}$ sent from factor nodes $\factor{\clique}$ to variable nodes $X_i$,
the domain of variable $X_i$ is divided into different pieces by constraints in formula $\formula_{\clique}$ that correspond to different integration bounds and thus the resulting messages from integration is again univariate piecewise integration.
This concludes the proof.
\end{proof}

\subsection{PROPOSITION~\ref{prop: messages and unnormalized distribution and WMI}}
\begin{proof}{(Proposition~\ref{prop: messages and unnormalized distribution and WMI})}
Given the tree structure of the factor graph as well as the factorization of WMI as in Equation~\ref{eq: factorized joint distribuiton},
the factors functions can be partitioned into groups, with each group associated with each factor nodes $\factor{\clique}$ that is a neighbour of the variable node $X_i$.
Then the unnormalized joint distribution can be rewritten as follows.
\begin{align*}
    p(\x) = \prod_{\factor{\clique} \in \neigh(X_i)} F_{\clique}(x_i, \x_\clique)
\end{align*}
where $\x_\clique$ denotes the set of all variables in the subtree connected to the variable $X_i$ via the factor node $\factor{\clique}$,
and $F_{\clique}(x_i, \x_\clique)$ denotes the product of all the factors in the group associated with factor $\factor{\clique}$.
Then we have that
\begin{align*}
    p(x_i) 
    &= \prod_{\factor{\clique} \in \neigh(X_i)} \msg{\factor{\clique}}{X_i}{}(X_i) \\
    &= \prod_{\factor{\clique} \in \neigh(X_i)} \int F_\clique(x_i, \x_\clique) ~d \x_\clique
    = \int p(\x) ~d \x \backslash x_i
\end{align*}
where the last equality is obtained by interchanging the integration and product. Thus it holds that $p(x_i)$ obtained from the product of messages to variable node $X_i$ is the unnormalized marginal. The fact that the partition function of marginal $p(x_i)$ is the WMI of formula $\formula$ follows Proposition~\ref{prop: partition function equals to WMI}.
\end{proof}

\subsection{PROPOSITION~\ref{prop: nilpotent matrix order}}
\begin{proof}{(Proposition~\ref{prop: nilpotent matrix order})}
W.l.o.g, assume that both the chosen root node and leaf nodes are variable nodes.
Recall that the tree-height $h$ is the longest path from root node to any leaf node.
Let $n_f$ be the number of factor nodes in the longest path in the factor graph from root node to a leaf node that defines the tree-height $h$.
Then it holds that $h = 2 n_f$ since the factor graph is a bipartite graph.

For another, consider a directed graph $\graph$ whose nodes are the directed factor nodes in $\mathcal{F}$ and whose directed edges go from one factor node to factor nodes if they are visited right after in the MP-WMI. By definition, we have that $A = 2c \cdot M$ where $M$ is the adjacency matrix of $\graph$, and $c$ is the constant that bounds the size of sub-formulas associated to factors.

For adjacency matrix $M$, since the power matrix $M^k$ has non-zero entries only when there exists at least one path in graph $\graph$ with length $k$,
the order of matrix $M$ is the length of longest path in graph $\graph$ plus one which is two times the number of number of factor nodes in the longest path in the factor graph, i.e., $2 n_f$. 
Therefore the adjacency matrix $M$ is a nilpotent matrix with order being at most $2 n_f$, i.e., the tree-height of the factor graph, which is at most the diameter of the factor graph. So is matrix $A$.
\end{proof}

\subsection{PROPOSITION~\ref{prop: upper bound of msg pieces}}
\begin{proof}{(Proposition~\ref{prop: upper bound of msg pieces})}
The statement $(i)$ holds since the message $\msg{X_i}{\factor{ij}}{}$ is the product of messages hence intersection of corresponding pieces by definition in Proposition~\ref{pro: recursive messages}.

For the statement $(ii)$, the end points of the message pieces in message $\msg{\factor{ij}}{X_j}{}$ are obtained by the solving linear equations with respect to variable $x_j$ as described in~\citet{zeng2019efficient} where they define them as critical points.
For these equations, each side can be either an endpoint in message $\msg{X_i}{\factor{ij}}{}$ or an $\lra$ atom from a literal in sub-formula $\formula_{ij}$.
Then there are at most $2mc$ equations with one side as an endpoint and the other size as an $\lra$ atom, and at most $c^2$ equations with both sides as $\lra$ atoms.
Thus the total number of critical points from solving the equations is $2mc + c^2$,
which indicates that the number of pieces, whose domains are bounded intervals with critical points being their endpoints, is at most $2mc + c^2$.
\end{proof}

\subsection{PROPOSITION~\ref{prop: recursive upper bound for msg pieces}}
\begin{proof}{(Proposition~\ref{prop: recursive upper bound for msg pieces})}
The proof is done by mathematical induction at steps in MP-WMI.
Given a directed factor node $\factor{s} \in \mathcal{F}$, denote the set $\mathcal{S}(\factor{s}) := \{ \factor{s^\prime} \mid A_{\factor{s}, \factor{s^\prime}} \neq 0 \}$.

For step $0$, the statement holds by the definition of $v^{(0)}$.
Suppose that for step $t$, each entry in vector $v^{(t-1)}$ denoted by $\overrightarrow{\factor{s}}$ bounds the number of pieces in the message $\msg{X_i}{\factor{s}}{}$ received by factor $\factor{s}$ from some variable node $X_i$ at step $t - 1$.
For step $t$, 
it holds for $v^{(t)}$ by its definition that 
$(v^{(t)})_{\factor{s}} = \sum_{\factor{s^\prime} \in \mathcal{S}(\factor{s})} (A_{\factor{s}, \factor{s^\prime}} (v^{(t-1)})_{\factor{s^\prime}} + c^2 )$.

Moreover, for a factor node $\factor{s} \in \mathcal{F}$, there exists an variable $X_i$ such that nodes in $\mathcal{S}(\factor{s})$ are connected to $\factor{s}$ by the variable node $X_i$ in the factor graph.
Since the entry $(v^{(t-1)})_{\factor{s^\prime}}$ bounds the number of message pieces in $\msg{X_j}{\factor{s^\prime}}{}$ for some variable $X_j$,
the number of message pieces in each message $\msg{\factor{s^\prime}}{X_i}{}$ is bounded by $2c \cdot (v^{(t-1)})_{\factor{s^\prime}} + c^2$ by Proposition~\ref{prop: upper bound of msg pieces}.
It further indicates that the number of message pieces in $\msg{X_i}{\factor{s}}{}$ is bounded by $\sum_{\factor{s^\prime} \in \mathcal{S}(\factor{s})} (2c \cdot (v^{(t-1)})_{\factor{s^\prime}} + c^2) = (v^{(t)})_{\factor{s}}$ since the non-zero entries in $A$ are defined as $2c$.
Thus the statement holds for step $t$, which finishes the induction and the proof.
\end{proof}

\subsection{PROPOSITION~\ref{prop: bounds for number of message pieces}}
\begin{proof}{(Proposition~\ref{prop: bounds for number of message pieces})}
For brevity, we denote the $L1$-norm by $\parallel \cdot \parallel$.
Denote the cardinality of set $\F$ to be $s$.
From the definition of matrix $A$, it holds that $\parallel A \parallel \leq 2 c s$.
Then for all $t$, it holds that 
\begin{equation*}
    \parallel v^{(t)} \parallel
    \leq \parallel A v^{(t-1)} + c^2 \cdot \sgn(A v^{(t-1)}) \parallel
    \leq 2 c s \parallel v^{(t-1)} \parallel + c^2 s
\end{equation*}
From the recurrence above, it can be obtained that
\begin{align*}
    &\parallel \sum_{t = 0}^{d} v^{(t)} \parallel
    \leq \sum_{t = 0}^{d} \parallel v^{(t)} \parallel \\
    &\leq \sum_{t = 0}^{d} [ (2 c s)^t \parallel v^{(0)} \parallel + \sum_{i=0}^{t-1} (2 c s)^i c s] \leq 2(2 c s)^{2d+2}
\end{align*}
Moreover, since the cardinality $s \leq 2 n$, we have that $\parallel~\sum_{t = 0}^{d} v^{(t)}~\parallel$ is of $\bigO((4 n c)^{2d + 2})$.
\end{proof}

\end{document}